# LEARNING TO DESIGN FROM HUMANS: IMITATING HUMAN DESIGNERS THROUGH DEEP LEARNING

**Ayush Raina**
Department of Mechanical Engineering
Carnegie Mellon University
Pittsburgh, PA 15213 USA
araina@andrew.cmu.edu

**Christopher McComb**
School of Engineering Design, Technology, and Professional Programs
The Pennsylvania State University
University Park, PA, 16802 USA
mccomb@psu.edu

**Jonathan Cagan**
Department of Mechanical Engineering
Carnegie Mellon University
Pittsburgh, PA 15213 USA
cagan@cmu.edu

## ABSTRACT

*Humans as designers have quite versatile problem-solving strategies. Computer agents on the other hand can access large scale computational resources to solve certain design problems. Hence, if agents can learn from human behavior, a synergetic human-agent problem solving team can be created. This paper presents an approach to extract human design strategies and implicit rules, purely from historical human data, and use that for design generation. A two-step framework that learns to imitate human design strategies from observation is proposed and implemented. This framework makes use of deep learning constructs to learn to generate designs without any explicit information about objective and performance metrics. The framework is designed to interact with the problem through a visual interface as humans did when solving the problem. It is trained to imitate a set of human designers by observing their design state sequences without inducing problem-specific modelling bias or extra information about the problem. Furthermore, an end-to-end agent is developed that uses this deep learning framework as its core in conjunction with image processing to map pixel-to-design moves as a mechanism to generate designs. Finally, the designs generated by a computational team of these agents are then compared to actual human data for teams solving a truss design problem. Results demonstrates that these agents are able to create feasible and efficient truss designs without guidance, showing that this methodology allows agents to learn effective design strategies.*

## INTRODUCTION

Advancements in machine learning and computational modelling have brought us closer to the goal of developing intelligent agents that can solve a variety of problems. Machine intelligence has surpassed human levels in several problem-solving milestones [1–5]. One common feature among all these problems is that they are well-defined with a set of rules and actions that are universally followed. Humans are still seen to perform better than machines for tasks which require skills like strategic reasoning, abstract decision making, creativity, and explainability. All of these skills are characteristic of the field of design. In design, the sheer complexity of most problems and the loosely-bound definitions and requirements necessitate a combination of the above-mentioned skills to generate solutions. This motivates the need to combine the diverse qualities of both humans and machines to facilitate solving challenges in design. This work focuses on developing methods to learn essential design strategies from human data and use that to help improve the performance of computational design agents.

Design is an iterative process; in order to create something, humans interact with an environment by making sequential decisions. Expert designers apply efficient search strategies to navigate massive design spaces [6]. The ability to navigate maze-like design problem spaces [7,8] by making relevant decisions is of great importance and is a crucial part of learning to emulate human design behavior. Based on this understanding, Figure 1 schematically represents a human solving a design problem, in this case a truss design problem as used in the example study in this paper. The problem is decomposed into two steps: perception and problem solving. The designer perceives the current design state and takes an action to modify it based on previous knowledge and understanding of the problem. This new design is then fed back into the current design state and the process is repeated. Designer behavior guides the decisions regarding what actions to take and is shown as a black box, the input and output of which are observable. However, what happens inside is hidden and not entirely known.

The motivation for this research is primarily based on understanding what goes on within that black box and uncovering the implicit rules and strategies. We present a data-driven approach that learns human design and problem-solving strategies and also learns to generate new designs only from observing the visual states of an evolving design, without explicitly being provided with any information about the performance metrics or meaning behind design modifications. Since the presented deep learning framework does not encode any specific information about allowable actions and instead represents the whole process using images, this research has promise to provide a domain-agnostic method of representing a generative design process. Every design process has a different set of actions or operations, preventing a common algorithm or framework to be used across problems. Our domain-agnostic method uses an image-based common representation to define and track the evolution of design problems.

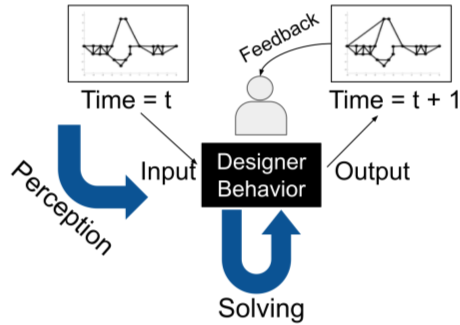

*Figure 1: Schematic representation of a human interacting with a design interface*

This work illustrates how sequential images from historical designer data can be used to emulate designer. Specifically, an independent agent algorithm is developed that utilizes methods from unsupervised representation learning, imitation learning and image processing to model the entire design process as shown in Figure 1. The agent uses a convolutional autoencoder to map the design states to low dimensional embeddings, then it uses a neural network-based transition function to predict a design embedding based on the current state and then maps the embedding back to the original image dimensions. This process produces an image of how the next design should look like. Finally, a control algorithm driven by image processing constructs outputs the parameters to execute the operation. This agent is trained offline, doesn't require carefully labelled state-action data and is agnostic to objectives and other parameters. Even though the current framework utilizes images (a *2*-dimensional array), the framework can be extended to other problems where the raw state is representable using an *N* dimensional array. The contributions made by this research can be listed as follows:

a) A generalized methodology is presented that can be used across various design problems to learn implicit design strategies organically from data, using a pixel-based representation.

b) An agent framework is developed that generates and follows design suggestions and automates the design process in an end-to-end manner.

c) The agent maintains a process similar to a human designer and hence provides a platform to use agent in computational teams or along with humans in a hybrid AI/human team

For the entirety of the paper the term *framework* is used to refer to the combined deep leaning model and the term *agent* is used to refer to the end-to-end algorithm that is the deep learning framework and the automated inference algorithm both combined The paper is organized as follows: Section 1 contains the review of the relevant studies and builds up a foundation for the different methods used in the framework. Section 2 begins by explaining the setup of the design study and the nature of the datasets used, then it explains the network architectures, training strategies and some intermediate results with the (image-based, or imaginal) deep learning part. Section 3 explains the experimental setup that was required to evaluate the performance of the agents and their computational teams. The results and discussions from the experiments are found in Section 4. The final conclusions and future work are mentioned at the end in Section 5.

## 1. BACKGROUND:

This section is divided into four subsections, each representing a different area of research this paper relates to. A review is conducted to discuss ideas and show how this work is related to the existing research in each of these fields and in what ways it is new and different from them.

### 1.1 Design as a sequential decision-making process:

The idea of problem solving (or design) as a sequential decision-making process is bolstered by similar representations used in robotics [9], game AI research[10], behavioral economics [11], decision-making [12], and also design research [13–15]. Here, the basic idea is that an agent is placed in a design environment and the goal of the agent is to interact and create a design maximizing the given performance metrics. Formalizing design as a problem-solving process opens up possibilities of applying logic, inference and different mathematical models to often abstractly-defined design processes. Modern-day design processes work in a computer-in-the-loop setting, where designers leverage computational tools along with their learned knowledge to make informed decisions [16]. Data collected from these processes can be utilized to extract key features and insights about them, while some of the previous studies have found that certain probabilistic models can be used to represent design strategies and describe skilled behavior [14,15], generate designs with similar trends to humans [17] and finally generalize them to transfer across certain problems [18]. Modeling design processes in an environment is challenging since, limiting the designer to a small set of operations can limit designer skills and these operations further have multiple parameters associated with them. Also, these variables and actions are specific to the design problem that is being modelled. This makes it difficult to develop an agent or a framework that can be used across different domain of problems since a new learning architecture is required every time. This problem is discussed and solved in the next subsections.

**1.2 Pixel based representation of design processes**

Humans are shown to think and solve problems visually [19], creating mental imagery [20] and incrementally making changes to it until reaching a solution [21]. Humans are able to abstract and store information as images and then recall them when solving a problem; this process is collectively called imaginal thinking. Some previous works have also shown its utility in helping humans solve design problems using an interface [22,23]. This motivates the use of an image (pixel based) common representation for defining a design problem. It is especially beneficial for design due to the great diversity of design variables in different problems leading to algorithms to be problem specific. An empty image encompasses the complete design space since any design can be represented within that image (if the problem is 2D), hence a design process can be represented as a designer changing the pixels of an image to achieve the best configuration of pixels. This form of representation can be extended to any problem that shows the current problem state as a set of pixels (or even voxels for 3D data). This method leads to a very high dimensional problem that can be very difficult to manipulate directly (e.g., for design). The research presented in the subsequent sections utilize image-based representations and formulates a learning problem based on it.

**1.3 Using deep learning for design feature extraction – dimensionality reduction:**

Dimensionality reduction of design spaces is a critical problem and numerous techniques have been discussed in the literature. Such reduction is done in order to use computation techniques efficiently, however there exists a tradeoff between geometric variability (possibility of creating novel designs) and dimensionality (complexity) that must be addressed [24]. Techniques like Karhunen-Loève expansion [10,11] and principal component analysis [12] are linear methods that determine a set of dimensions that can maximally explain the variance in the data. Higher accuracy non-linear methods have also been utilized by researchers like kernel PCA and multi-dimensional scaling [25,26] however these methods are non-invertible, i.e., one can go from high dimensional space to a low dimension embedding, however, the embedding cannot be converted to the original design. This is an essential requirement of our framework. Recent advances in deep learning has led to development of methods for highly non-linear dimension reduction that are invertible and accurate. Autoencoders are one such family of methods which can determine lower dimensional embeddings [25,27–30] and is an efficient data-driven way of learning representations [31]. Autoencoders are a type of neural network that are trained to reconstruct an output that is identical to the input. Often a density constraint limits the dimensions of the embedding and forces the autoencoder to generate a lower dimensional representation. Autoencoders have proven to perform much better than other available methods, especially for design tasks [25,32]. A variant of these autoencoders, a deep convolutional autoencoder, combines a convolutional architecture [33] with the encoder-decoder structure of an autoencoder [30]. It has the ability to extract high-level semantic information from image data and encode it to a lower dimension embedding. It is used in this paper as a part of a larger framework to encode design information to a lower dimensional embedding and back to an image which is human readable. Adding further constraints on the distribution of the embedding elements leads to a variational autoencoder (VAE) architecture [34], which has certain advantages in design generation tasks. However, that requires higher amount of data hence it may be explored in future works. Other recent works in design have also used convolution deep learning architectures to solve different tasks with both 2D and 3D data with interesting results [35–37].

**1.4 Mimicry – learning to imitate from raw pixels:**

Understanding what happens inside the human behavior "black box" has been approached differently in different fields of research. Historically, researchers from psychology consider learning human behavior as *learning a function* that generates the same output given a specific input. These methods were aimed at learning rule-based relationships [38,39], or learning similarities [40,41] in problem solving behavior. Studies in cognitive science have put forth several core ingredients of human intelligence, including intuitive physics [42–45], problem decomposition skills [46,47], ability in learning-to-learn [48], and others [49]. More recently learning from imitation or *imitation learning* has been actively utilized to solve various problems since it doesn't require significant state-space exploration, making it a faster approach in practice. Researchers have taught robots to imitate and learn how to drive [50], reach [51], manipulate [52], play Atari games [53] and also fly and control a helicopter [54]. However, all these require properly labelled state-action pairs, and such datasets are very few in number and very problem-specific. Recent work by Liu et. al [55], proposes a framework that learns only by observing state-space trajectories. This allows a network to learn in third person (or use historical data) which is closer to how humans learn by just observing without explicit action information. The research presented builds off some of the ideas from that work and adapts it to work with design data. Some other works develop on the two step problem representation idea by having an encoded representation of the environment and a control network that can navigate it [56], however they do not utilize any human knowledge and use active learning methods like deep reinforcement learning to solve the problem, which is different from our approach.

Significant research has been done in deriving qualitative explanations for human design strategies [8,57–60] and using them to further design optimization algorithms [61,62]. Also, methods have been developed to extract and verify design knowledge from design data, like inter part dependencies from design databases [63], learning a machine learning based design method recommender [64] and also developing more efficient optimization algorithms [65] from crowdsourced design solutions. These previous efforts, however, do not address the problem of making a generalized parameter-independent algorithm and solving a sequential decision-making design problem which is an essential factor for our framework in order to make it similar to how humans solve it. In conclusion, prior research relates to the methods that we've used; however, our work integrates concepts from different fields and proposes a unique two step deep learning framework that decomposes the solving process of design similar to a human, and hence makes it compatible for future hybrid teaming possibilities.

## 2. METHODS: A Deep Learning Framework for Visual Design Evolution

In this section the methods involved in this research are explained. The first subsection discusses the original design study and the source of the dataset. The second subsection is a formulation of the problem. The third subsection explains the model architecture and the training strategies used. The final subsection illustrates how the deep learning framework is integrated with an inference algorithm to develop an agent.

### 2.1 Design study and database

The nature of the developed framework is such that it can utilize historical data for training and is generalizable to any problem with a pixel-based state representation. For the purpose of the work addressed in this paper the database used for learning comes from a truss design study (conducted by McComb et al [66]). The study involved senior undergraduate students in mechanical engineering who were tasked to design a truss structure given a certain set of objectives to fulfil. The students were randomly divided in sixteen teams (three members each). An interface was designed for the study which allowed students to apply one of the nine operations shown in Figure 2 and also interact with the rest of the team members to complete the truss structure. Two metrics were provided to participants for feedback on the quality of these trusses: factor of safety (FOS) and total mass. The former is a metric to represent how strong and resistant to failure the design is, where a higher value represents higher strength. A value above 1 illustrates sufficient strength and deems the design feasible. Mass shows the amount of material that has been used in the structure. These two metrics together are used to evaluate the strength to weight ratio (SWR) i.e. the ratio of the FOS and mass. This represents how optimized the structure is and is also used to evaluate the quality of a structure; a high FOS with low mass will lead to a high SWR showing that high strength was achieved with minimal mass. During the study, the design interface was also used to capture data whenever a subject made any change in the design. The study was conducted in a controlled environment where different constraints were provided to the subjects in gradual time steps.

The resulting database contains the design state information for all the subjects under different design scenarios. Previous work used this sequential design data to develop a method of representing design strategies and contrast high and low performing designers [14,15,17]. For the purpose of our research we are using only the first portion of the study which is unconstrained. The first image in Figure 2 shows the initial state of the design problem, with the pink arrows representing the load bearing nodes while the other nodes are the support nodes. The database contains only the design state information with nodes and members, excluding any loading arrows or water. The subjects were tasked to create feasible designs (>1 FOS) with minimal mass of the truss structure. This consists of a total of 12850 design states that appear in design sequences generated by humans. These states include parametric specifications of nodes and members that exist in the design structure. As the designs progress, the number of nodes and members vary greatly in the dataset, changing the number of variables to completely define a design. In the current work, the design state is represented by an image with a fixed number of pixels instead of the variable number of parametric values. This representation leads to a more visually interpretable representation which is what the subjects in the study actually observed. Every design in the solution space can be fully represented with an image. Most real-life design problems can be visualized similarly using software tools which provide state-space information to the human users along with a list of actions to modify/create the design. Decomposing design as a sequential decision-making process simplifies it to the schematic shown in Figure 1. Having an image as the primary representation of a state can allow developing a framework that can be used across multiple problems since there is no problem specific modeling in terms of actions or design variables. Even though, the truss design problem is visually simpler in comparison to real-life mechanical design problems, it still captures the basic challenges of perception followed by selecting actions and relevant parameters. The focus of the current work is on developing this method for truss design problem, however extending to other problems is a tangible next step and the effectiveness shall be tested and discussed in future work.

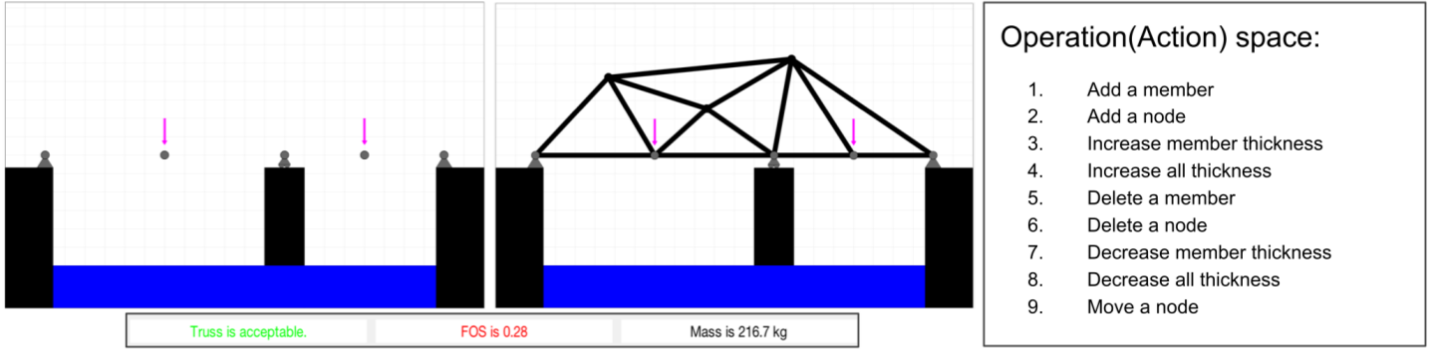

*Figure 2: Depicting the interface for the truss design software along with the visible metric and possible operations [66].*

**2.2 Problem Formulation**

The problem that this work deals with is formulated as a prediction problem, where the framework is expected to predict the next design state given previous state information. The dataset used contains a set of sequence demonstrations $\{D_1, D_2, ... D_n\}$ of human generated designs from the truss study. Each demonstration is sequence of state observation $D_i = [\ o^i_1, o^i_2, ... o^i_T\ ]$ where, $o^i_t$ is the observation at time (analogous to iteration number, since one iteration is one time unit) $t$ in the $i^{th}$ demonstration sequence and is an image that shows the current design (truss structure). Each designer creates multiple demonstrations and all demonstrations begin from the same initial state. Every demonstration is a trajectory through different intermediate design states that eventually leads to a final design, representing a search through the design space. This dataset does not contain any parameters for the metrics for node and members or information about the loading parameters. However, there are implicit relationships between consecutive states corresponding to design operations which lead to changes in pixel intensities. The framework (in Section 3.3) aims to learn those implicit relationships among these states from the dataset since they correspond to designer strategies for the given problem and boundary conditions.

This task to replicate human strategies can be formulated as learning a function that converts $o^i_t$ to $o^i_{t+1}$ ( $\forall\ t + 1 < T_i$ ; where $T_i$ is the total time for the $i^{th}$ demonstration). In order to develop new designs in the study humans were provided with a set of nine operations (A) that change the existing designs; the set A = $\{a_1, a_2, ... , a_9\ \}$, and each of the operations have a parameter set associated with them (w). Hence, $o^i_{t+1} = a_j(o^i_t, w)$ represents generating a new design by applying operation $j$ with parameters $w$. The whole design process can be seen as a designer analyzing the current state $o^i_t$, selecting and applying one of these operation functions with relevant parameters, and then repeating the process all over again. The designer looks at the current state and makes the decision about the operations and the parameters using implicit design knowledge and is able to eventually create a well performing truss structure. The database captures that information as it preserves the sequence of the progression of design. The desired framework must satisfy the following requirements:

a) Ability to semantically interpret the design features of state $o^i_t$.

b) Create an embedding that can represent large variability in the designs with less features.

c) Map consecutive design states to learn how design operations function.

d) Learn how designs progress iteratively to generate design trajectories.

The first two requirements relate to the perception part of the problem and the next two relate to problem solving. In order to fulfil them two different neural network architectures are used. They are explained in the next section.

**2.3 Model framework:**

To learn the low dimensional embeddings of the design data, a convolutional auto-encoder is designed. It has two parts, encoder (*e(.)*) and decoder (*d(.)*). The encoder is used to encode the raw pixel data into a semantically meaningful embedding that can capture all the important features of the structure at a reduced size. The decoder on the other hand is used to convert this low dimensional embedding back into the original size image. This step helps visually realize the image. Another neural network is used as a transition network (*t(.)*), that takes the embedding from the encoder as the input and converts it into a new design embedding that helps in progressing the design, and which is regained by the decoder. Figure 3 shows the complete deep learning framework.

**2.3.1 Perception - dimension reduction:**

This part of the framework deals with perception of the design. The raw designs are high-dimensional and using them to predict other high dimensional designs is computationally intensive. Thus, identifying semantically meaningful embeddings that are low dimensional will make later operations less computationally intensive. This was achieved with a nine-layer autoencoder (as shown in Figure 4).

Dimension reduction was achieved in the encoder by using a combination of 3 layers of stride 2 convolutions [67] and a max-pool layer. The kernel sizes for the convolution layers were in decreasing order (12x12), (9x9) and (5,5) with increasing number of channels- 32, 64 and 128, respectively, in order to capture large features in the first layer (like nodes and member thicknesses). For the decoder, 1 un-pooling layer and 3 de-convolution (or convolution transpose [68]) layers with stride 2 were used for up-sampling the dimensions. The kernel sizes were in increasing order (7,7), (9,9) and (12,12). The number of channels were same as the encoder but in a reverse order (128, 64, 32). Non-linear activations were used for all the layers, all convolution layers except for the final layer used ReLU activations [69] because of their efficiency in deep-networks, while the linear layer at the center and the final convolutional layers uses *sigmoid* activation to avoid grey value and forcing the network to make a decision for every pixel (since *sigmoid* is biased towards 0 or 1 values). The dataset was split into test and train images with a 80-20 split (80% training, 20% test data) and the autoencoder was trained in an un-supervised manner since there are no labels and the network automatically identifies the optimal latent representations. The main objective for training an autoencoder is to reconstruct the same input, so that the output is the same as the input. Also, an information bottleneck is designed in the hidden layers to reduce the dimensions. The auto-encoder reduces the initial 76x76 image (5776 dimensions) down to 512 dimensions in the embedding. A mean square error (MSE) was used as the loss function to train the autoencoder. Adam optimizer [70] was used to train the weights of the network. A sufficient low value of 0.0012 MSE was achieved with a binary accuracy of over 91% over the test set. An R-squared value of 0.9684 is achieved in the test case, indicating that the network is capable of explaining nearly all of the variance in the data.

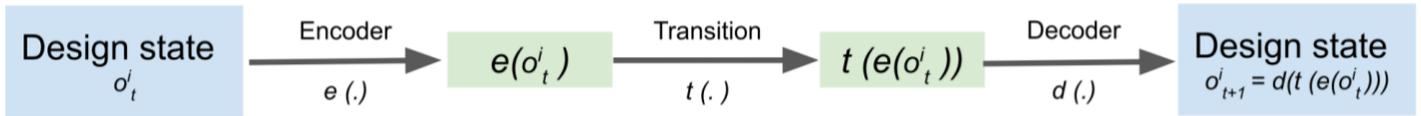

*Figure 3: A block (functional) diagram of the deep learning framework*

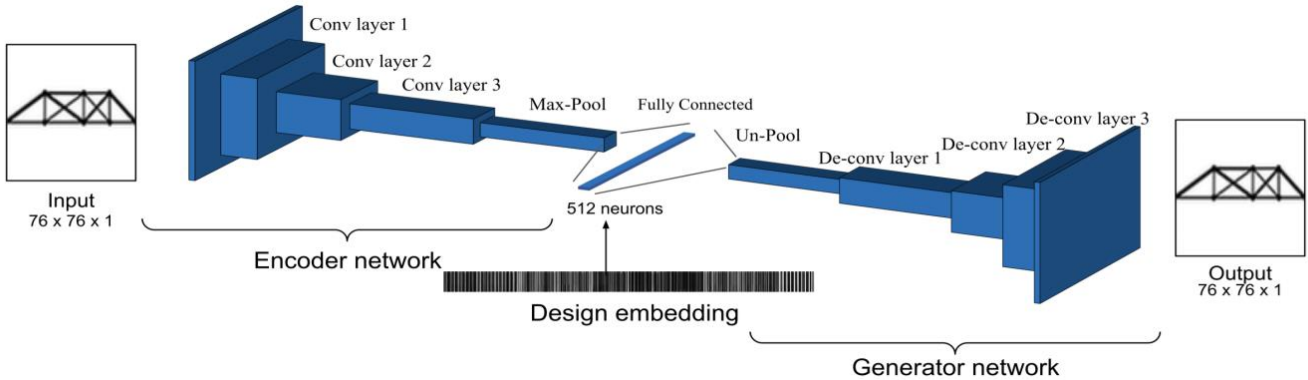

*Figure 4: Network architecture for the convolutional autoencoder*

**2.3.2 Prediction - the transition network:**

This part of the framework is a non-linear function that maps the embedded representations of designs in consecutive time steps. This function basically manipulates the current designs to generate a new design that shall eventually lead to a complete final design. This manipulation of the current designs to generate a new design is similar to how a human designer approaches a design problem, in a step-by-step manner taking feedback and applying relevant operations. A neural network is used to approximate this function. The network takes 5 previous design embeddings as input (5 x 512 units), concatenates them and then down-samples it in the two linear layers (1024, 512 units) back to 512 units. A non-linear activation function Leaky-ReLU [71] was used for all the linear layers. Multiple design states are concatenated in order to make sequential information accessible to the network for better predictions [3]. The weights of the neural network implicitly represent the design operations that are applied to manipulate the design states. These operations modify the design embeddings by adding, deleting or manipulating their values (matrix operations) in order to generate new designs. Since this step manipulates an abstract representation of a visual image it is similar to how humans build designs [21].

This network was trained after the autoencoder. Training data for the transition network consist of embedding sequences from the design study. $S = \{ e(o^i_t), e(o^i_t), ... e(o^i_t)\} \forall i \in H$ *(all sequences)* is the total set of embedding sequences, for the purpose of this network with 5 inputs, the dataset was converted to $S = \{ [e(o^i_1), e(o^i_2), ... e(o^i_5); e(o^i_6)], [e(o^i_2), e(o^i_3), ... e(o^i_6); e(o^i_7)], ....[e(o^i_{Ti-5}), e(o^i_{Ti-4}), ... e(o^i_{Ti-1}); e(o^i_{Ti})] \} \forall i \in H$. The semi-colon separates the input data from the labels i.e. the next design state in the sequence. For training, test and training sequences were separated with a 80-20 split (80% training data, 20% test) and the network was evaluated on how well it performed on the test set. This converts the imitation problem into a supervised learning problem where based on the previous 5 design embeddings the next embedding needs to be predicted. The training data used is only the state sequences from the design study, so the network is only trying to learn the implicit relationships between the states; it has no explicit understanding of design goals, constraints or intent. The transition network was combined with the encoder and decoder networks as shown in Figure 5 to complete the deep learning framework. For the training of the transition network weights for the encoder-decoder networks were kept fixed and MSE loss was calculated. Adam optimizer [70] was then used to backpropagate the loss values to train the weights.

The training was accomplished in a 2-step process where the first step pre-trains the network and the second step fine tunes to predict meaningful designs. In the first step, the network must learn to recreate the current design itself. This pre-training step is important since the weights of the network are initialized randomly and training may not lead to meaningful predictions [72]; pre-training helps in finding better weights by providing good initialization to the network weights. In this fine-tuning step the learning rate is lowered, and the network is trained to recreate the next design in the sequence. Through experimentation the mentioned architecture was finalized for the transition network, a final MSE of 0.0072 (or binary accuracy of 90.05%) is achieved for the test case and an R-squared value of 0.8105 was achieved by the final network. This indicates that the network was able to explain a large majority of the variance in the data. The final network was arrived at through an iterative hyper-parameter search, where numerous architectures were compared, preference was given to smaller networks as larger networks tend to over-fit the data. A major boost in performance was seen when past design states were concatenated in the input, showing the importance of the relation between design trajectories and next state prediction.

**2.3.3 Framework results**

After training both the autoencoder and the transition network, updated weights were put together in the overall framework as shown in Figure 5. The input side of the framework consists of 5 instances of the encoder networks. The embeddings generated by those networks are then concatenated together and a 3-layer neural network down-samples the embeddings before the decoder network and converts the embedding back to the full image size. The final generated image from the framework within an iteration is the suggestion how the next design should look, hence creating a mental image of the design solution. However, these are still pixel-based information and specific parameters need to be identified to create an actual design.

To visualize the suggested move better, a difference of the generated and input image is taken and then overlaid as a heatmap on the original image. Figure 6 shows 3 images: the current design, the suggestion heatmap and the ground truth (i.e., the actual subsequent design from data). The suggestion heatmap shows the prediction of design change as a color gradient on the current design state. The color gradient spans across pink to green with 0 (no change) values being represented as black. Pink regions represent the regions where the network adds materials in the prediction while at the green regions it deletes or reduces the material. Black implies no change is predicted in those regions. The ground truth is the next design state from the original dataset, showing the actual design created by the designer in the study. A heatmap that matches the ground truth shows perfect prediction. Figure 7 shows some more sequences that have been taken from the test set which contain designs that were not seen in the training phase. From both Figure 6 and Figure 7 it can be observed that the network successfully learns to predict semantically meaningful pixel regions (that appear like discrete operations relating to nodes and members) and also correspond to the operations undertaken by the human designers in the test dataset. This shows that the framework was able to extract design features from the problem space along with learning to developing truss designs to match human operations using supervised imitation learning. However, it can be seen that there is some level of noise in the heatmaps which can be improved in future work by using more advanced architectures and other methods of training. Currently the framework is able to provide exceedingly well results using very basic architectures, utilizing more complex architectures and methods may be needed for more complicated visual setting (more than nodes and lines) and also may lead to better results in terms of noise and efficiency however, for the scope of current work the framework appears to be sufficient. The network is also seen to produce several suggestions at once, showing that it is unsure about which move to take. That may be because of several reasons: at a given design state there may not be a single best operation to take that mimics a human strategy; or different sequences of operations can be used to reach the same final state hence which operation to do first could be arbitrary; or at a particular state the design can go in various possible trajectories and if the network is unsure it produces more noise and less discrete features; also the network could possibly be learning an 'average' of multiple designers and their strategies, since it is not trained to only follow a particular style of design generation (although that would be feasible with the current network). In future work, further analysis needs to be done that can identify the primary factor for multiple operation suggestions. Since the heatmaps do seem to follow a singular 'design idea' and the suggestions are coherent, the current performance is sufficient and promising for it to act as a design suggestion engine. Now, in order to evaluate how well the suggestions perform in an actual design process, an inference algorithm (similar to having a control system) needs to be designed that can map these suggestion heatmaps (ideas) to the operator parameters (actions) and then select and implement a single operation from the list required to sequentially generate a parametric truss design.

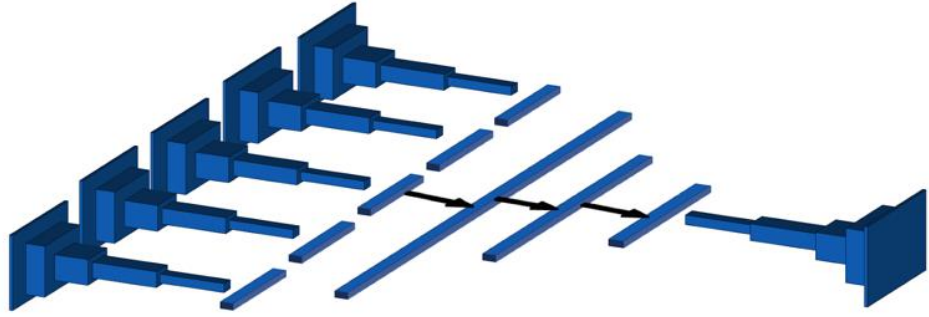

*Figure 5: Network architecture for the complete framework*

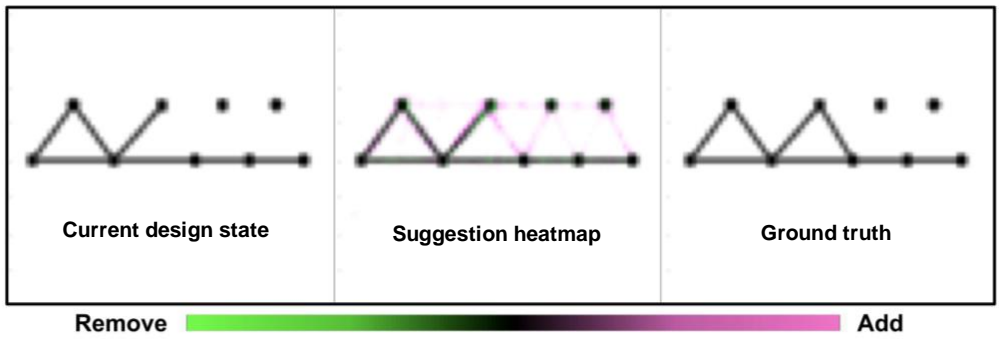

*Figure 6: Suggestion heatmaps from the DL framework. The color gradient shows the suggestions where pink means "Add", green means "Delete" and black means no change*

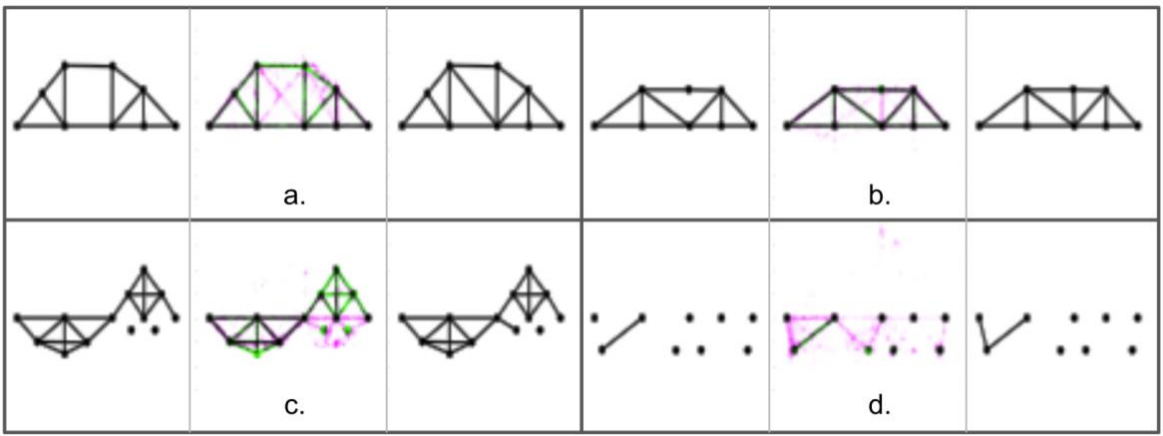

*Figure 7: (a-d) represent different test dataset pairs and the heatmaps from the best performing framework.*

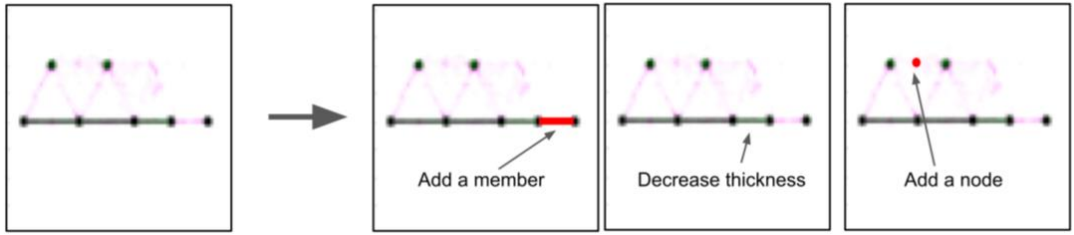

*Figure 8: An example of a candidate list generation with labelled operations generated in one of the applications of the image processing algorithm.*

## 2.4 Developing a deep learning agent

The framework generates the design as a picture and the suggestions/changes are shown as the highlighted pixels in the heatmaps. These images are basically how the deep learning framework imagines/suggests that the design should progress in the next few steps and is a pictorial visualization with no detailed information about which parts are members, their size, or where the nodes are. However, in order to evaluate a design, it needs to have parametric information for the nodes and members involved in the change/suggestion. Interpreting this heatmap information to make changes to the current design is a trivial activity for humans; however, it is very complex to map the process algorithmically. A rule-based algorithm was developed to act like the control system of the framework which can automatically inference from the heatmaps and map pixel-based suggestions to operator parameters. When combined with the framework, this completes the process of generation of a new design state given the current design states. Details of the rule-based inference algorithm are explained in the Appendix.

In brief, the algorithm uses image processing techniques to pre- process and isolate different regions that may correspond to distinct operations from the heatmap. Further, the regions are analyzed based on the different parameters to check and classify it as a particular operation and finally a candidate list is prepared based on the different regions; these candidate lists can appear similar to Figure 8. The final step in the algorithm is to select one operation from the list to finally apply on the design. This is achieved by comparing the similarity between the suggestion and the candidates and the one with the highest similarity is probabilistically selected. Hence, the algorithm goes through these different steps and outputs the particular operation parameters to change the design of the truss structure. The framework is shown in Figure 9, it can be considered an independent agent as it senses an environment, makes decisions, and then acts on the environment generating new designs [73]. The design state is the sequence of the previous 5 designs which is used as an input for the framework. This engine generates a suggestion heatmap, which is fed into the inference algorithm. This algorithm processes the suggestions and classifies the pixel representation into a design operation. This move is then executed, and a new design is generated.

## 3. EXPERIMENTAL SETUP

This section details how the experiments were carried out to evaluate the quality of the framework by creating a team of identical deep learning-based agents. In order to evaluate the performance of the agent algorithm we compare the generated designs with human data from the original design study. The design study was conducted in a team (of 3 designers) and they interacted at predefined intervals. When interacting, members of a team shared their designs with the option for each designer to select any one of the 3 designs in the team to continue working on (typically the best design of the triplet). In order to make a fair comparison, 3 different agent instances of the algorithm were combined into teams and they were made to interact periodically at the same average rate as the humans, at which point the best design from the agent team is accepted by all agents. Every agent is given a random starting point and then iteratively refines their design. Each agent works independently to generate new designs that are fed back into the deep learning framework to complete the loop. The agents in a team interact every $48^{th}$ move (mean value from data), when the interaction occurs the current design from every agent is shared in a common pool and the highest performing feasible design (based on SWR and FOS) is adopted by all the agents. When the agents don't interact then the design states of individual agents are updated only with their own designs. This greedy method of interaction was chosen as a simple approximation to the process of human interaction in a team, which is complicated and modeling that is beyond the scope of the current work. This is the only step where the team of agents are implicitly aware of the metrics. However, this step is separate from the design generation steps of the agents (heatmap generation and operation selection) and occurs in < 2% operations; hence the agents individually still continue to work in a metric agnostic manner for generating trusses. The complete loop is shown in Figure 9 and is repeated for 250 iterations per agent (equivalent to the human average). This whole setup was repeated to produce 16 team solutions in order to compare with the 16 design teams in the study.

## 4. RESULTS AND DISCUSSION

Data from the experiments, generated by deep learning agents, and human designers is analyzed in this section. Figure 10, Figure 11 and Figure 12 show the plots for performance metrics: Factor of Safety (FOS) and Strength to Weight Ratio (SWR) with respect to number of iterations. The lines in the background connect the data-points (designs) from the same designer showing trajectories of how the design progressed. The comparison is made between blue (solid) and green (dashed) trajectories where blue color (solid) represents the Deep Learning Agents (DLAgent) and the green color (dashed) is for Human data. The plot presents the data for all 48 humans and 48 DLAgents. The graph limits to only 250 iterations for the purpose of better visualization since the majority of the data can be shown within this range; for calculating the performance values all iterations are considered. Due to the nature of truss design, whenever a new node is added it is virtually hanging in the air and hence the overall structure is infeasible. This leads to zero values of FOS and SWR, which is not a good representation of the quality of the truss structure since, minus that extra node, the structure could be feasible. This leads to significant amount of noise in the data where the quality metrics unnecessarily dive to zero. In order to avoid that, a denoising operation is carried out once a design becomes acceptable in the process, the most recent non-zero value is adopted as the current value. This way the data is denoised while also maintaining the trends. The solid blue and green lines show the mean values at the given iteration for the respective class. Mean values are shown in order to provide a sense of the average trend in the data. Figure 10 shows a plot of the FOS over time. The red dotted line shows the threshold for a feasible design (FOS = 1); all designs above that line are feasible. It can be observed that although a few human designers reach high FOS values very early in the process, most of the humans struggle to create a feasible design as they remain below the red line, showing that there is a mix of designers throughout. On the other hand, DLAgent builds slowly but more consistently since the majority of the designs beyond ~175 iterations reach levels higher than humans. Also, the mean lines further justify the performance by DLAgents since there is a gradual but consistent increase in FOS and it exceeds the human level and eventually the feasible level. Also, for the human data there is some noise in the mean with immediate ups and downs which smooths out with increased iterations, but there is less noise with the agents. This could be indicative of learning/searching behavior for humans since they had access to immediate metric feedback and could try different things and evaluate their design whereas agents solve the problem in mostly a metric agnostic manner.

Figure 11 shows the plot for comparing the SWR over time. Humans appear to perform better than the DLAgents especially in the early part of the process since the dashed green mean line is higher than solid blue. Select individual human designers also seem to have reached much higher values throughout the process. However, it should be noted that a high SWR can be achieved with infeasible FOS (< 1.0) but very low mass values. This motivated us to calculate a refined SWR value as shown in Figure 12.

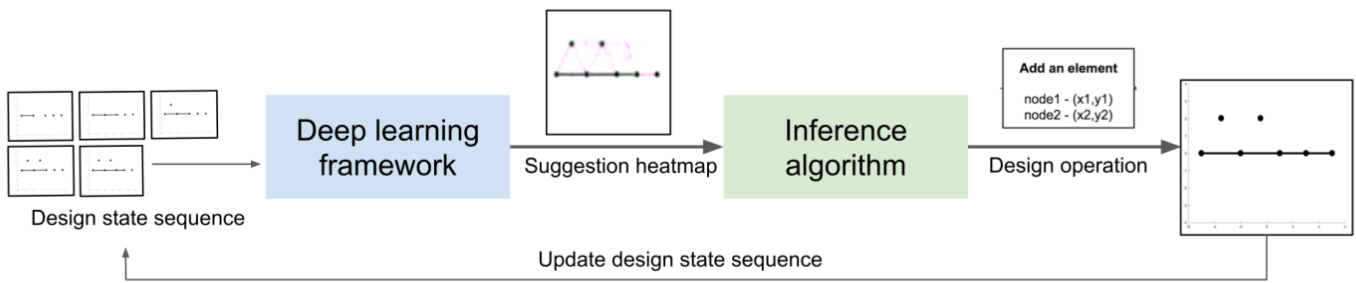

*Figure 9: Schematic representation of the complete agent*

For Figure 12, only designs that are feasible (with FOS ≥ 1.0) are allocated a SWR value unlike in Figure 11 where an infeasible design could also contribute in representing the quality of the design. Here a majority of the human data-points are removed, showing that most of the designs in the early part of the process with high SWR were actually incomplete or infeasible designs. We observe that even though designs by DLAgents don't reach very high values like some over-performing humans, there still are a considerable number of designs that are feasible and have good SWR, and the agents are much more consistent with the quality of their designs. Observing the mean lines, DLAgents seem to perform quite close to the humans throughout and slightly exceed them in the later part of the process. The gradual increase in SWR throughout the process shows that the deep learning framework was able to learn certain design strategies from human data and is able to implement them in the generation process.

Figure 13 shows the mean value across the best designs created by individual designers for the two performance metrics. It shows that the DLAgent performs better in terms of FOS and comes considerably close in terms of SWR. This implies that the DLAgents have learnt to mimic certain aspects of the human strategies and learn to generate design sequences even after they started from mere random initial points. It must be noted that the DLAgents are generating designs only on the basis of transition features learnt from data. On the other hand, the humans had access to various metrics like FOS, mass, and also a colored grading showing which members were under stress at every iteration. All of this information guided them to produce better designs in the process. Agents receive no information about mass, and hence cannot get direct feedback for how heavy the design is which makes limiting it difficult, possibly leading to achieving higher mass in their structures. This helps the designs with strength (FOS) but adversely affects the SWR. However, as seen later in the SWR result plots the agents maintain good SWR values meaning that the agents learn to efficiently utilize the extra mass. It can be concluded that the agents have learned to create feasible good performing designs, however, humans were still more efficient in the early parts of the process indicating a room for improvement.

## 5. CONCLUSION

Human designers possess great problem-solving skills. Essential insights can be extracted from them to develop computer agents that can work with humans in unison as a team. A novel methodology is presented in this paper that is used to develop agents that learn purely from human data and generate high performing designs. The methodology involves training a deep learning framework on historical human design data just by observation. The framework learns to generate designs without any specific design operation information. The generated design is basically a visualization (imagination) of how the new design should look like. Then, that design is fed into a rule-based algorithm that acts as the control system of the agent. Its task is to apply design operations to realize the visualization. This work integrates methods from deep learning, concepts from psychology and behavior modelling to achieve design automation in a comprehensive end-to-end manner. A two-step deep learning framework is developed that can perceive the design state information from pixels and then generate new designs after learning to imitate human designers. Finally, a rule-based inference algorithm converts the image into a parametric truss structure which allows the algorithm to act like an agent as it can now iteratively work on its own to generate complete designs in a manner similar to humans.

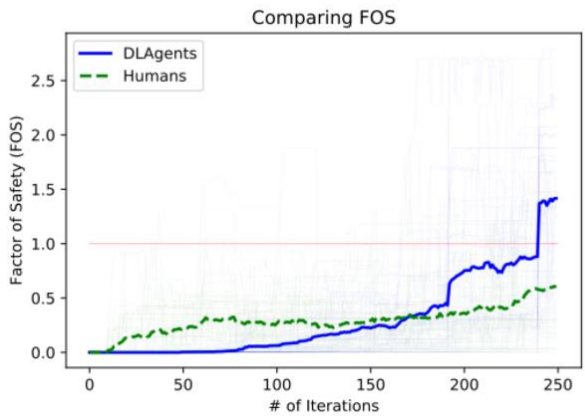

*Figure 10 Comparing DLAgents and Humans on FOS*

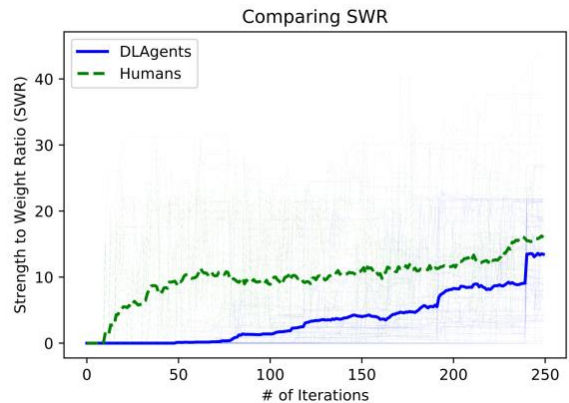

*Figure 11 Comparing DLAgents and Humans on SWR*

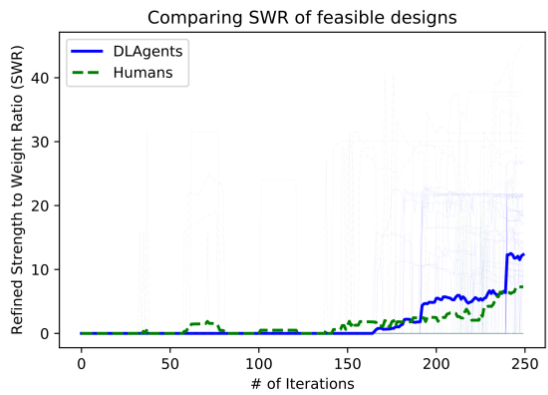 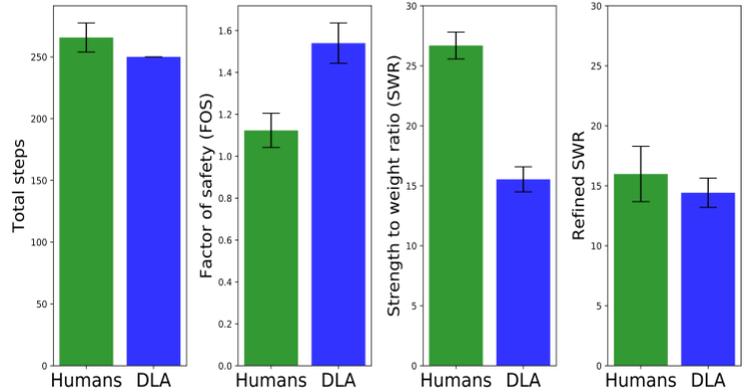

*Figure 12 Comparing DLAgents and Humans on RSWR*     *Figure 13: Compiled results. Error bars represent the standard error.*

The comparative results show that the agent learns to create designs that are comparable to humans designs from the study. While making this comparison it must also be considered that the agent did not have access to real-time feedback on the quality of the design or the objectives of the process (only the team interaction was implicitly guided by it). Humans, on the other hand, could get feedback at every step helping them in the process. Moreover, the agents start the design process from randomly generated initial states, while humans begin on their own. However, irrespective of the differences the agents were seen to perform exceedingly well and achieve feasible designs that are even comparable to human level. This implies that the agents were able to extract implicit design strategies from data and use them to generate designs. The framework has the ability to learn the relationships between consecutive design states and hence can be applied to consecutive decision-making design processes whose sequential states have incremental changes and can be represented as an *N* dimensional array.

Learning from historical human data helps the agent identify the important regions of a design space and implicitly learn to navigate it, helping reach good performing designs. This imitation learning, however, has its limitations since the agent can at best do as well as the teacher, i.e., the human in this case, and also it may only learn to work with designs similar to shown in the training data in case of overfitting. In order to overcome this, the networks are trained separately to first create design embeddings that can learn general features of a truss design and hence can represent even new unseen truss designs and then learn generic design operations on them. This can allow the agents to explore new designs by possible interpolations between design embeddings. The agent might still learn bad strategies in case the quality of the subject is bad. The aim for the current research was to illustrate an ability to extract design strategies from data however in future work knowledge about metrics could be infused to ignore low performing strategies along with careful selection of training data. Further experimentation also needs to be carried out to test the novelty in designs and how different they are from the training data to evaluate generality of the learnt operations. This motivates us to utilize and test other methods of learning in the agents as well as identify the effect of training data on the final performance. For the current study the agents were only trained to imitate humans using metric agnostic design data. In the future, embedding these agents in a goal driven environment where real-time rewards are provided for optimal designs can lead to interesting results. Also, creating hybrid approaches of imitation learning with other methods of active learning can also be explored since it can significantly enhance the agent performance.

**ACKNOWLEDGEMENTS**

This paper has been submitted for publication in the 2019 Design Automation Conference (IDETC). This material is based upon work supported by the Defense Advanced Research Projects Agency through cooperative agreement No. N66001-17-1-4064. Any opinions, findings, and conclusions or recommendations expressed in this paper are those of the authors and do not necessarily reflect the views of the sponsors.

**REFERENCES**

[1]  Campbell, M., Hoane, A. J., and Hsu, F., 2002, "Deep Blue," Artificial Intelligence, 134(1), pp. 57–83.

[2]  Mnih, V., Kavukcuoglu, K., Silver, D., Rusu, A. A., Veness, J., Bellemare, M. G., Graves, A., Riedmiller, M., Fidjeland, A. K., Ostrovski, G., Petersen, S., Beattie, C., Sadik, A., Antonoglou, I., King, H., Kumaran, D., Wierstra, D., Legg, S., and Hassabis, D., 2015, "Human-Level Control through Deep Reinforcement Learning," Nature, 518, p. 529.

[3]  Silver, D., Huang, A., Maddison, C. J., Guez, A., Sifre, L., van den Driessche, G., Schrittwieser, J., Antonoglou, I., Panneershelvam, V., Lanctot, M., Dieleman, S., Grewe, D., Nham, J., Kalchbrenner, N., Sutskever, I., Lillicrap, T., Leach, M., Kavukcuoglu, K., Graepel, T., and Hassabis, D., 2016, "Mastering the Game of Go with Deep Neural Networks and Tree Search," Nature, 529, p. 484.

[4]  Brown, N., and Sandholm, T., 2018, "Superhuman AI for Heads-up No-Limit Poker: Libratus Beats Top Professionals," Science, 359(6374), pp. 418 LP – 424.

[5]  Vinyals, O., Babuschkin, I., Chung, J., Mathieu, M., Jaderberg, M., Czarnecki, W. M., Dudzik, A., Huang, A., Georgiev, P., Powell, R., Ewalds, T., Horgan, D., Kroiss, M., Danihelka, I., Agapiou, J., Oh, J., Dalibard, V., Choi, D., Sifre, L., Sulsky, Y., Vezhnevets, S., Molloy, J., Cai, T., Budden, D., Paine, T., Gulcehre, C., Wang, Z., Pfaff, T., Pohlen, T., Yogatama, D., Cohen, J., McKinney, K., Smith, O., Schaul, T., Lillicrap, T., Apps, C., Kavukcuoglu, K., Hassabis, D., and Silver, D., 2019, *AlphaStar: Mastering the Real-Time Strategy Game StarCraft II*.

[6]  Cross, N., 2004, "Expertise in Design: An Overview," Design Studies, 25(5), pp. 427–441.

[7]  Newell, A., and Simon, H. A., 1972, *Human Problem Solving*, Prentice-Hall, Inc., Upper Saddle River, NJ, USA.

[8]  Daly, S., McKilligan, S., Christian, J., Seifert, C., and Gonzalez, R., 2012, "Design Heuristics in Engineering Concept Generation," Journal of Engineering Education, 101.

[9]  Ross, S., 2013, "Interactive Learning for Sequential Decisions and Predictions," PhD Thesis, Carnegie Mellon University.

[10] Yannakakis, G. N., and Togelius, J., 2018, *Artificial Intelligence and Games*, Springer Publishing Company, Incorporated.

[11] Payne, J. W., Bettman, J. R., and Johnson, E. J., 1993, *The Adaptive Decision Maker.*, Cambridge University Press, New York, NY, US.

[12] Busemeyer, J. R., and Townsend, J. T., 1993, "Decision Field Theory: A Dynamic-Cognitive Approach to Decision Making in an Uncertain Environment.," Psychol Rev, 100(3), pp. 432–459.

[13] SINGER, D. J., DOERRY, N., and BUCKLEY, M. E., 2009, "What Is Set-Based Design?," Naval Engineers Journal, 121(4), pp. 31–43.

[14] McComb, C., Cagan, J., and Kotovsky, K., 2017, "Capturing Human Sequence-Learning Abilities in Configuration Design Tasks Through Markov Chains," Journal of Mechanical Design, 139(9), pp. 91101–91112.

[15] McComb, C., Cagan, J., and Kotovsky, K., 2017, "Mining Process Heuristics From Designer Action Data via Hidden Markov Models," Journal of Mechanical Design, 139(11), p. 111412.

[16] Finger, S., and R. Dixon, J., 1989, "A Review of Research in Mechanical Engineering Design. Part II: Representations, Analysis, and Design for the Life Cycle," Research in Engineering Design, 1, pp. 121–137.

[17] McComb, C., Cagan, J., and Kotovsky, K., 2017, "Utilizing Markov Chains to Understand Operation Sequencing in Design Tasks," *Design Computing and Cognition '16*, pp. 401–418.

[18] Raina, A., McComb, C., and Cagan, J., 2018, "Design Strategy Transfer in Cognitively-Inspired Agents," *44th Design Automation Conference*.

[19] Brooks, R. A., 1991, "New Approaches to Robotics.," Science, 253(5025), pp. 1227–1232.

[20] Athavankar, Uday A . 1997, "Mental Imagery As A Design Tool," *Cybernetics and Systems*, 28(1), pp. 25–42.

[21] Goldschmidt, Gabriela 1992, "Serial Sketching: Visual Problem Solving In Designing -," *Cybernetics and Systems,* 23(2), pp. 191–219.

[22] Yin, Y. H., Xie, J. Y., Xu, L. D., and Chen, H., 2012, "Imaginal Thinking-Based Human-Machine Design Methodology for the Configuration of Reconfigurable Machine Tools," IEEE Transactions on Industrial Informatics, 8(3), pp. 659–668.

[23] Yin, Y. H., Zhou, C., and Zhu, J. Y., 2010, "A Pipe Route Design Methodology by Imitating Human Imaginal Thinking," CIRP Annals, 59(1), pp. 167–170.

[24] Diez, M., Campana, E. F., and Stern, F., 2015, "Design-Space Dimensionality Reduction in Shape Optimization by Karhunen–Loève Expansion," Computer Methods in Applied Mechanics and Engineering, 283, pp. 1525–1544.

[25] D'Agostino, D., Serani, A., Campana, E. F., and Diez, M., 2018, "Nonlinear Methods for Design-Space Dimensionality Reduction in Shape Optimization BT - Machine Learning, Optimization, and Big Data," G. Nicosia, P. Pardalos, G. Giuffrida, and R. Umeton, eds., Springer International Publishing, Cham, pp. 121–132.

[26] Chen, W., Fuge, M., and Chazan, J., 2017, "Design Manifolds Capture the Intrinsic Complexity and Dimension of Design Spaces," Journal of Mechanical Design, 139(5), pp. 51102–51110.

[27] Yumer, M. E., Asente, P., Mech, R., and Kara, L. B., 2015, "Procedural Modeling Using Autoencoder Networks," *Proceedings of the 28th Annual ACM Symposium on User Interface Software & Technology*, ACM, New York, NY, USA, pp. 109–118.

[28] Guo, T., Lohan, D., Cang, R., Ren, Y., and Allison, J., 2018, *An Indirect Design Representation for Topology Optimization Using Variational Autoencoder and Style Transfer*.

[29] D'Agostino, D., Serani, A., Campana, E., and Diez, M., 2018, *Deep Autoencoder for Off-Line Design-Space Dimensionality Reduction in Shape Optimization*.

[30] Hinton, G. E., and Salakhutdinov, R. R., 2006, "Reducing the Dimensionality of Data with Neural Networks," Science, 313(5786), p. 504.

[31] Bengio, Y., Courville, A. C., and Vincent, P., 2012, "Unsupervised Feature Learning and Deep Learning: {A} Review and New Perspectives," CoRR, abs/1206.5538.

[32] McComb, C., 2018, *Towards the Rapid Design of Engineered Systems Through Deep Neural Networks*.

[33] LeCun, Y., Bottou, L., and Haffner, P., 1998, "Gradient-Based Learning Applied to Document Recognition."

[34] Kingma, D. P., and Welling, M., 2013, "Auto-Encoding Variational Bayes," eprint arXiv:1312.6114, p. arXiv:1312.6114.

[35] Zhang, Y., Chen, A., Peng, B., Zhou, X., and Wang, D., 2019, "A Deep Convolutional Neural Network for Topology Optimization with Strong Generalization Ability," arXiv:1901.07761 [cs, stat].

[36] Banga, S., Gehani, H., Bhilare, S., Patel, S., and Kara, L., 2018, "3D Topology Optimization Using Convolutional Neural Networks," arXiv:1808.07440 [physics, stat].

[37] Burnap, A., Liu, Y., Pan, Y., Lee, H., Gonzalez, R., and Papalambros, P. Y., 2016, "Estimating and Exploring the Product Form Design Space Using Deep Generative Models," p. V02AT03A013.

[38] Carroll, J. D., 1963, "Functional Learning: The Learning Of Continuous Functional Mappings Relating Stimulus And Response Continua," *ETS Research Bulletin Series*, 1963(2), pp. i–144.

[39] Koh, K., and Meyer, D. E., 1991, "Function Learning: Induction of Continuous Stimulus-Response Relations.," J Exp Psychol Learn Mem Cogn, 17(5), pp. 811–836.

[40] DeLosh, E. L., Busemeyer, J. R., and McDaniel, M. A., 1997, "Extrapolation: The Sine qua Non for Abstraction in Function Learning.," J Exp Psychol Learn Mem Cogn, 23(4), pp. 968–986.

[41] Busemeyer, J. R., Byun, E., Delosh, E. L., and McDaniel, M. A., 1997, "Learning Functional Relations Based on Experience with Input-Output Pairs by Humans and Artificial Neural Networks.," Knowledge, concepts and categories., pp. 408–437.

[42] S. Spelke, E., Gutheil, G., and Van de Walle, G., 2019, *The Development of Object Perception.*

[43] Baillargeon, R., Li, J., Ng, W., and Yuan, S., 2008, *An Account of Infants' Physical Reasoning*.

[44] Bates, C., Yildirim, I., B Tenenbaum, J., and W Battaglia, P., 2015, *Humans Predict Liquid Dynamics Using Probabilistic Simulation*.

[45] Gershman, S. J., Horvitz, E. J., and Tenenbaum, J. B., 2015, "Computational Rationality: A Converging Paradigm for Intelligence in Brains, Minds, and Machines," Science, 349(6245), pp. 273 LP – 278.

[46] Kulkarni, T. D., Narasimhan, K., Saeedi, A., and Tenenbaum, J. B., 2016, "Hierarchical Deep Reinforcement Learning: Integrating Temporal Abstraction and Intrinsic Motivation," CoRR, abs/1604.06057.

[47] Biederman, I., 1987, "Recognition-by-Components: A Theory of Human Image Understanding.," Psychological Review, 94(2), pp. 115–147.

[48] Thrun, S., and Pratt, L., 1998, *Learning to Learn: Introduction and Overview*.

[49] Lake, B. M., Ullman, T. D., Tenenbaum, J. B., and Gershman, S. J., 2016, "Building Machines That Learn and Think Like People," CoRR, abs/1604.00289.

[50] Pomerleau, D. A., 1989, "ALVINN: An Autonomous Land Vehicle in a Neural Network," *Advances in Neural Information Processing Systems 1*, D.S. Touretzky, ed., Morgan-Kaufmann, pp. 305–313.

[51] Billard, A., and Matarić, M. J., 2001, "Learning Human Arm Movements by Imitation:: Evaluation of a Biologically Inspired Connectionist Architecture," Robotics and Autonomous Systems, 37(2), pp. 145–160.

[52] Finn, C., Yu, T., Zhang, T., Abbeel, P., and Levine, S., 2017, "One-Shot Visual Imitation Learning via Meta-Learning," CoRR, abs/1709.04905.

[53] Hester, T., Vecerik, M., Pietquin, O., Lanctot, M., Schaul, T., Piot, B., Sendonaris, A., Dulac-Arnold, G., Osband, I., Agapiou, J., Leibo, J. Z., and Gruslys, A., 2017, "Learning from Demonstrations for Real World Reinforcement Learning," CoRR, abs/1704.03732.

[54] Abbeel, P., Coates, A., and Ng, A. Y., 2010, "Autonomous Helicopter Aerobatics through Apprenticeship Learning," The International Journal of Robotics Research, 29(13), pp. 1608–1639.

[55] Liu, Y., Gupta, A., Abbeel, P., and Levine, S., 2017, "Imitation from Observation: Learning to Imitate Behaviors from Raw Video via Context Translation," CoRR, abs/1707.03374.

[56] Ha, D., and Schmidhuber, J., 2018, "World Models," CoRR, abs/1803.10122.

[57] Pretz, J. E., 2008, "Intuition versus Analysis: Strategy and Experience in Complex Everyday Problem Solving.," Memory & cognition, 36(3), pp. 554–566.

[58] Cagan, J., Dinar, M., J. Shah, J., Leifer, L., Linsey, J., Smith, S., and Vargas Hernandez, N., 2013, *Empirical Studies of Design Thinking: Past, Present, Future*.

[59] Björklund, T. A., 2013, "Initial Mental Representations of Design Problems: Differences between Experts and Novices," Design Studies, 34(2), pp. 135–160.

[60] Egan, P., and Cagan, J., 2016, "Human and Computational Approaches for Design Problem-Solving BT - Experimental Design Research: Approaches, Perspectives, Applications," P. Cash, T. Stanković, and M. Štorga, eds., Springer International Publishing, Cham, pp. 187–205.

[61] Cagan, J., and Kotovsky, K., 1997, "Simulated Annealing and the Generation of the Objective Function: A Model of Learning During Problem Solving," Computational Intelligence, 13(4), pp. 534–581.

[62] McComb, C., Cagan, J., and Kotovsky, K., 2018, *Drawing Inspiration From Human Design Teams For Better Search And Optimization: The Heterogeneous Simulated Annealing Teams Algorithm*.

[63] Matthews, P., Blessing, L., and Wallace, K. M., 2002, *The Introduction of a Design Heuristics Extraction Method*.

[64] Fuge, M., Peters, B., and Agogino, A., 2014, *Machine Learning Algorithms for Recommending Design Methods*.

[65] Sexton, T., and Ren, M. Y., 2017, "Learning an Optimization Algorithm Through Human Design Iterations," Journal of Mechanical Design, 139(10), pp. 101404–101410.

[66] McComb, C., Cagan, J., and Kotovsky, K., 2018, "Data on the Design of Truss Structures by Teams of Engineering Students," Data Brief, 18, pp. 160–163.

[67] Springenberg, J. T., Dosovitskiy, A., Brox, T., and Riedmiller, M. A., 2014, "Striving for Simplicity: The All Convolutional Net," CoRR, abs/1412.6806.

[68] Fergus, R., Zeiler, M. D., Taylor, G. W., and Krishnan, D., 2010, "Deconvolutional Networks," *2010 IEEE Computer Society Conference on Computer Vision and Pattern Recognition(CVPR)*, pp. 2528–2535.

[69] Nair, V., and E. Hinton, G., 2010, *Rectified Linear Units Improve Restricted Boltzmann Machines Vinod Nair*.

[70] Kingma, D. P., and Ba, J., 2014, "Adam: A Method for Stochastic Optimization," CoRR, abs/1412.6980.

[71] Maas, A. L., Hannun, A. Y., and Ng, A. Y., 2013, "Rectifier Nonlinearities Improve Neural Network Acoustic Models," *In ICML Workshop on Deep Learning for Audio, Speech and Language Processing*.

[72] Bengio, Y., Lamblin, P., Popovici, D., and Larochelle, H., 2006, "Greedy Layer-Wise Training of Deep Networks," *Proceedings of the 19th International Conference on Neural Information Processing Systems*, MIT Press, Cambridge, MA, USA, pp. 153–160.

[73] Franklin, S., and Graesser, A., 1997, "Is It an Agent, or Just a Program?: A Taxonomy for Autonomous Agents," *Intelligent Agents III Agent Theories, Architectures, and Languages*, J.P. Müller, M.J. Wooldridge, and N.R. Jennings, eds., Springer Berlin Heidelberg, pp. 21–35.

[74] Tarjan, R., 1971, "Depth-First Search and Linear Graph Algorithms," *12th Annual Symposium on Switching and Automata Theory (Swat 1971)*, pp. 114–121.

[75] Rodriguez, J., and Ayala, D., *Erosion and Dilation on 2D and 3D Digital Images: A New Size-Independent Approach*.

[76] Zhou Wang, A. C. Bovik, H. R. Sheikh, and E. P. Simoncelli, 2004, "Image Quality Assessment: From Error Visibility to Structural Similarity," IEEE Transactions on Image Processing, 13(4), pp. 600–612.

# APPENDIX

1. Details of the rule-based image-processing algorithm:

The algorithm is developed to infer operation parameters from suggestion heatmaps generated by the deep learning framework. This algorithm works as a control system of the agents which helps them realize the visualizations of the next design, generated by the deep learning framework. The rules are specific to the design problem at hand and would need to be modified for application across problems. However, the deep learning framework can remain consistent. The algorithm provides different methods for identifying moves related to nodes and members. They are detailed as follows (also shown in Figure 14). through three basic steps: a pre-processing step, a classification step and, finally, the selection and application of operation.

1.1 Preprocessing and isolating the suggestions:

Node: The preprocessing step involves thresholding the heatmap so that node regions can be clearly identified and be isolated from noisy pixels. For isolating the suggestions, a *counting islands algorithm* (a depth first search algorithm [74]) is used to identify pixel regions with high intensity to isolate multiple possible nodes and each region is further analyzed in the next part of the algorithm to determine its candidacy.

Member: For detecting members it is essential to preserve the structure/shape of the highlighted pixels. Hence a combination of dilation and erosion algorithm [75] is used to enhance the suggestion features/shape and then a threshold is applied to get rid of the extra noise. Applying these processing techniques have shown to be useful while extracting meaningful features from an image [75] and showed improvements in inference results for the current work.

1.2 Classify the Operation(s):

Node: Based on the pixel intensities, center of mass is computed for each of the isolated regions from the previous step. The location of this point defines the parameters associated with this node operation. Finally, feasibility conditions are checked to ensure that it is possible to apply the operation. Operations that are associated with nodes are "Add a Node" and "Delete a Node".

Member: For classifying member-based operations, every pair of nodes is evaluated since, in order to create a member, there must exist two nodes in the design. Based on the intensity and the number of the highlighted pixels in the region between a pair of nodes, that candidate pair is further evaluated for feasibility and a relevant operation is assigned. Operations associated with members are "Add Member", "Delete Member", Increase thickness" and "Decrease thickness".

1.3 Select and Apply the operation:

Once the heatmap is evaluated for all possible suggested operations, a candidate list is generated with the operation type and relevant parameters. Finally, the candidate image that is most similar to the framework generated design is probabilistically selected. This similarity is evaluated by comparing the image structure (SSIM index [76]). A higher SSIM value means that the selected candidate image is closer to what the deep learning framework suggested. This metric is, however, not a perfect method to identify the best candidate from the list since the similarity can be biased towards high pixel operations like adding members in comparison to adding a node which involves smaller pixel regions. In order to overcome that a probabilistic component is introduced in selection of the final operation from the candidates.

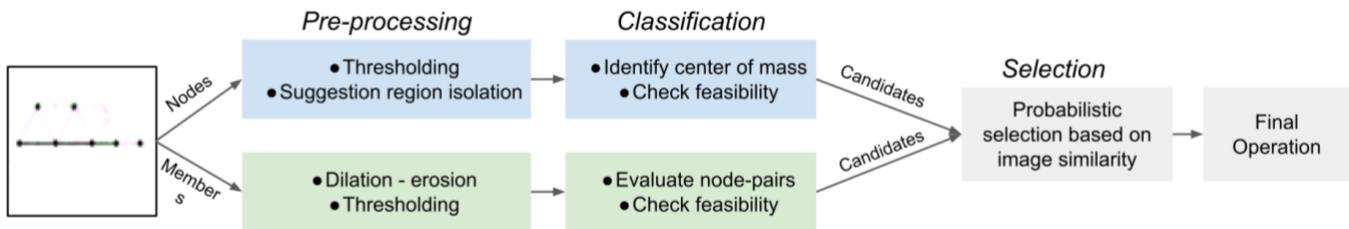

*Figure 14: Algorithm for rule-based inference automation*